\pdfoutput=1
\documentclass[11pt]{article}

\usepackage{acl}

\usepackage{times}
\usepackage{latexsym}
\usepackage[T1]{fontenc}
\usepackage[utf8]{inputenc}
\usepackage{microtype}
\usepackage{helvet}
\usepackage{amsmath}
\usepackage{amssymb}
\usepackage{courier}
\usepackage{bm}
\usepackage{algpseudocode}
\usepackage[noend,ruled,linesnumbered]{algorithm2e}
\usepackage{color}
\usepackage{enumitem}
\usepackage{url}
\usepackage{graphicx}
\usepackage{array}
\usepackage{multirow}
\usepackage{pifont}
\usepackage{xcolor,colortbl}
\usepackage{setspace}
\usepackage{subfigure}

\newcommand{\bme}{{\bm e}}
\newcommand{\bmu}{{\bm u}}
\newcommand{\bmv}{{\bm v}}
\newcommand{\bX}{{\bf X}}

\newcommand{\bU}{{\bf U}}
\newcommand{\bV}{{\bf V}}
\DeclareMathOperator*{\argmax}{arg\,max}

\newcommand{\cmark}{\ding{51}}
\newcommand{\xmark}{\ding{55}}
\definecolor{myred}{rgb}{0.85, 0.25, 0.25}
\definecolor{myblue}{rgb}{0.3, 0.3, 0.85}

\newcommand{\model}{SeeTopic}

\title{Seed-Guided Topic Discovery with Out-of-Vocabulary Seeds}

\author{Yu Zhang$^1$, Yu Meng$^1$, Xuan Wang$^1$, Sheng Wang$^2$, Jiawei Han$^1$ \\
  $^1$University of Illinois at Urbana-Champaign, IL, USA \\
  $^2$University of Washington, Seattle, WA, USA \\
  \texttt{\small \{yuz9,yumeng5,xwang174,hanj\}@illinois.edu\ \ \ \ swang@cs.washington.edu}}

\begin{document}
\maketitle

\begin{spacing}{0.97}
\begin{abstract}
Discovering latent topics from text corpora has been studied for decades. Many existing topic models adopt a fully unsupervised setting, and their discovered topics may not cater to users' particular interests due to their inability of leveraging user guidance. Although there exist seed-guided topic discovery approaches that leverage user-provided seeds to discover topic-representative terms, they are less concerned with two factors: (1) the existence of out-of-vocabulary seeds and (2) the power of pre-trained language models (PLMs). 
In this paper, we generalize the task of seed-guided topic discovery to allow out-of-vocabulary seeds. We propose a novel framework, named \textsc{\model}, wherein the general knowledge of PLMs and the local semantics learned from the input corpus can mutually benefit each other. 
Experiments on three real datasets from different domains demonstrate the effectiveness of \textsc{\model} in terms of topic coherence, accuracy, and diversity.\footnote{The code and datasets are available at \\ \url{https://github.com/yuzhimanhua/SeeTopic}.}
\end{abstract}

\section{Introduction}
Automatically discovering informative and coherent topics from massive text corpora is central to text analysis through helping users efficiently digest a large collection of documents \cite{griffiths2004finding} and advancing downstream applications such as summarization \cite{wang2009multi,wang2022textomics}, classification \cite{chen2015dataless,meng2020text}, and generation \cite{liu2021graphine}.

Unsupervised topic models have been the mainstream approach to topic discovery since the proposal of pLSA \cite{hofmann1999probabilistic} and LDA \cite{blei2003latent}. Despite their encouraging performance in finding informative latent topics, these topics may not reflect user preferences well, mainly due to their unsupervised nature. For example, given a collection of product reviews, a user may be specifically interested in product categories (e.g., ``\textit{books}'', ``\textit{electronics}''), but unsupervised topic models may generate topics containing different sentiments (e.g., ``\textit{good}'', ``\textit{bad}''). To consider users' interests and needs, seed-guided topic discovery approaches \cite{jagarlamudi2012incorporating,gallagher2017anchored,meng2020discriminative} have been proposed to find representative terms for each category based on user-provided seeds or category names.\footnote{In this paper, we use ``seeds'' and ``category names'' interchangeably.} However, there are still two less concerned factors in these approaches. 

\begin{table}[t]
\centering
\caption{Three datasets \cite{cohan2020specter,mcauley2013hidden,zhang2017react} from different domains and their topic categories (i.e., seeds). \textbf{\textcolor{myred}{Red}}: Seeds never seen in the corpus (i.e., out-of-vocabulary). In all three datasets, a large proportion of seeds are out-of-vocabulary.}
\scalebox{0.625}{
\begin{tabular}{>{\centering\arraybackslash}m{1.57cm}|cc}
\hline
\textbf{Dataset} & \multicolumn{2}{c}{\textbf{Category Names (Seeds)}}  \\ \hline
\textbf{SciDocs} (Scientific Papers) & \begin{tabular}[c]{@{}c@{}}cardiovascular diseases\\ chronic kidney disease\\ \textbf{\textcolor{myred}{chronic respiratory diseases}}\\ diabetes mellitus\\ \textbf{\textcolor{myred}{digestive diseases}}\\ \textbf{\textcolor{myred}{hiv/aids}}\end{tabular} & \begin{tabular}[c]{@{}c@{}}\textbf{\textcolor{myred}{hepatitis a/b/c/e}}\\ mental disorders\\ musculoskeletal disorders\\ \textbf{\textcolor{myred}{neoplasms (cancer)}} \\ neurological disorders\end{tabular} \\ \hline
\textbf{Amazon} (Product Reviews)  & \begin{tabular}[c]{@{}c@{}}\textbf{\textcolor{myred}{apps for android}} \\ books\\ \textbf{\textcolor{myred}{cds and vinyl}}\\ \textbf{\textcolor{myred}{clothing, shoes and jewelry}}\\ electronics\end{tabular}                                                & \begin{tabular}[c]{@{}c@{}}\textbf{\textcolor{myred}{health and personal care}}\\ \textbf{\textcolor{myred}{home and kitchen}}\\ movies and tv\\ \textbf{\textcolor{myred}{sports and outdoors}}\\ video games\end{tabular}                \\ \hline
\textbf{Twitter} (Social Media Posts) & \begin{tabular}[c]{@{}c@{}}food\\ \textbf{\textcolor{myred}{shop and service}}\\ \textbf{\textcolor{myred}{travel and transport}}\\ \textbf{\textcolor{myred}{college and university}}\\ \textbf{\textcolor{myred}{nightlife spot}}\end{tabular}                                            & \begin{tabular}[c]{@{}c@{}}residence\\ \textbf{\textcolor{myred}{outdoors and recreation}}\\ \textbf{\textcolor{myred}{arts and entertainment}}\\ \textbf{\textcolor{myred}{professional and other places}} \end{tabular}                   \\ \hline
\end{tabular}
}
\label{tab:intro}
\vspace{-0.5em}
\end{table}

\vspace{1mm}

\noindent \textbf{The Existence of Out-of-Vocabulary Seeds.} Previous studies \cite{jagarlamudi2012incorporating,gallagher2017anchored,meng2020discriminative} assume that all user-provided seeds must be \textbf{in-vocabulary} (i.e., appear at least once in the input corpus),
so that they can utilize the occurrence statistics or Skip-Gram embedding methods \cite{mikolov2013distributed} to model seed semantics. However, user-interested categories can have specific or composite descriptions, which may never appear in the corpus. 
Table \ref{tab:intro} shows three datasets from different domains: scientific papers, product reviews, and social media posts. In each dataset, documents can belong to one or more categories, and we list the category names provided by the dataset collectors. These seeds should reflect their particular interests. In all three datasets, we have a large proportion of seeds (45\% in SciDocs, 60\% in Amazon, and 78\% in Twitter) that never appear in the corpus. Some category names are too specific (e.g., ``\textit{chronic respiratory diseases}'', ``\textit{nightlife spot}'') to be exactly matched, others are the composition of multiple entities (e.g., ``\textit{hepatitis a/b/c/e}'', ``\textit{neoplasms (cancer)}'', ``\textit{clothing, shoes and jewelry}'').\footnote{One possible idea to deal with composite seeds is to split them into multiple seeds. However, there are many possible ways to express the conjunctions (e.g., ``\textit{/}'', ``\textit{()}'', ``\textit{,}'' and ``\textit{and}'' in Table \ref{tab:intro}), which may require manual tuning. Besides, simple chunking rules will induce splits that break the semantics of the original composition (e.g., ``\textit{professional and other places}'' may be split into ``\textit{professional}'' and ``\textit{other places}''). Moreover, even after the split, some seeds are still out-of-vocabulary. Therefore, we propose to use PLMs to tackle out-of-vocabulary seeds in a unified way. In experiments, we will show that our model is able to tackle composite seeds. For example, given the seed ``\textit{hepatitis a/b/c/e}'', we can find terms relevant to ``\textit{hepatitis b}'' and ``\textit{hepatitis c}'' (see Table \ref{tab:case}).}

\vspace{1mm}

\noindent \textbf{The Power of Pre-trained Language Models.} Techniques used in previous studies are mainly based on LDA variants \cite{jagarlamudi2012incorporating} or context-free embeddings \cite{meng2020discriminative}. Recently, pre-trained language models (PLMs) such as BERT~\cite{devlin2019bert} have achieved significant improvement in a wide range of text mining tasks. In topic discovery, the generic representation power of PLMs learned from web-scale corpora (e.g., Wikipedia or PubMed) may complement the information a model can obtain from the input corpus. Moreover, out-of-vocabulary seeds usually have meaningful in-vocabulary components (e.g., ``\textit{night}'' and ``\textit{life}'' in ``\textit{nightlife spot}'', ``\textit{health}'' and ``\textit{care}'' in ``\textit{health and personal care}''). The optimized tokenization strategy of PLMs \cite{sennrich2016neural,wu2016google} can help segment the seeds into such meaningful components (e.g., ``\textit{nightlife}'' $\rightarrow$ ``\textit{night}'' and ``\textit{\#\#life}''), and the contextualization power of PLMs can help infer the correct meaning of each component (e.g., ``\textit{\#\#life}'' and ``\textit{care}'') in the category name. Therefore, PLMs are much needed in handling out-of-vocabulary seeds and effectively learning their semantics.

\vspace{1mm}

\noindent \textbf{Contributions.} Being aware of these two factors, in this paper, we study seed-guided topic discovery in the presence of out-of-vocabulary seeds. Our proposed \textsc{\model} framework consists of two modules: 
(1) The \textit{general} representation module uses a PLM to derive the representation of each term (including out-of-vocabulary seeds) based on the general linguistic knowledge acquired through pre-training.
(2) The \emph{seed-guided local} representation module learns in-vocabulary term embeddings specific to the input corpus and the given seeds. 
In order to optimize the learned representations for topic coherence, which is commonly reflected by pointwise mutual information (PMI) \cite{newman2010automatic}, our objective implicitly maximizes the PMI between each word and its context, the documents it appears, as well as the category it belongs to. 
The learning of the two modules is connected through an iterative ensemble ranking process, in which the general knowledge of PLMs and the term representations specifically learned from the target corpus conditioned on the seeds can complement each other.

To summarize, this study makes three contributions. (1) \textit{Task}: we propose to study seed-guided topic discovery in the presence of out-of-vocabulary seeds. (2) \textit{Framework}: we design a unified framework that jointly models general knowledge through PLMs and local corpus statistics through embedding learning. (3) \textit{Experiment}: extensive experiments on three datasets demonstrate the effectiveness of \textsc{\model} in terms of topic coherence, accuracy, and diversity.

\section{Problem Definition}
As shown in Table \ref{tab:intro}, we assume a seed can be either a single word or a phrase. Given a corpus $\mathcal{D}$, we use $\mathcal{V}_{\mathcal{D}}$ to denote the set of \textbf{terms} appearing in $\mathcal{D}$. In accordance with the assumption of category names, each term can also be a single word or a phrase. In practice, given a raw corpus, one can use existing phrase chunking tools \cite{manning2014stanford,shang2018automated} to detect phrases in it. \textit{After} phrase chunking, if a category name is still not in $\mathcal{V}_{\mathcal{D}}$, we define it as \textbf{out-of-vocabulary}.

\vspace{1mm}

\noindent \textbf{Problem Definition.} 
\textit{Given a corpus $\mathcal{D}=\{d_1,...,d_{|\mathcal{D}|}\}$ and a set of category names $\mathcal{C}=\{c_1,...,c_{|\mathcal{C}|}\}$ where some category names are out-of-vocabulary, the task is to find a set of in-vocabulary terms $\mathcal{S}_i=\{w_{1},...,w_{S}\} \subseteq \mathcal{V}_{\mathcal{D}}$ for each category $c_i$ such that each term in $\mathcal{S}_i$ is semantically close to $c_i$ and far from other categories $c_j$ $(\forall j \neq i)$.}

\section{The \textsc{\model} Framework}
In this section, we first introduce how we model general and local text semantics using a PLM module and a seed-guided embedding learning module, respectively. Then, we present the iterative ensemble ranking process and our overall framework.

\subsection{Modeling General Text Semantics using a PLM}
PLMs such as BERT \cite{devlin2019bert} aim to learn generic language representations from web-scale corpora (e.g., Wikipedia or PubMed) that can be applied to a wide variety of text-related applications.
To transfer such general knowledge to our topic discovery task, we employ a PLM to encode each category name and each in-vocabulary term to a vector. To be specific, given a term $w \in \mathcal{C}\cup \mathcal{V}_{\mathcal{D}}$, we input the sequence ``\texttt{[CLS]} $w$ \texttt{[SEP]}'' into the PLM. 
Here, $w$ can be a phrase containing multiple words, and each word can be out of the PLM's vocabulary. To deal with this, most PLMs use a pre-trained tokenizer \cite{sennrich2016neural,wu2016google} to segment each unseen word into frequent subwords. 
Then, the contextualization power of PLMs will help infer the correct meaning of each word/subword, so as to provide a more precise representation of the whole category.

After LM encoding, following \cite{sia2020tired,thompson2020topic,li2020sentence}, we take the output of all tokens from the last layer and average them to get the term embedding $\bme_w$. In this way, even if a seed $c_i$ is \textbf{out-of-vocabulary}, we can still obtain its representation $\bme_{c_i}$.

\subsection{Modeling Local Text Semantics in the Input Corpus}
\label{sec:local}
The motivation of topic discovery is to discover latent topic structures from the input corpus. Therefore, purely relying on general knowledge in the PLM is insufficient because topic discovery results should adapt to the input corpus $\mathcal{D}$. Now, we introduce how we learn another set of embeddings $\{\bmu_w|w\in \mathcal{V}_{\mathcal{D}}\}$ from $\mathcal{D}$. 

Previous studies on embedding learning assume that the semantic of a term is similar to its local context \cite{mikolov2013distributed}, the document it appears \cite{tang2015pte,xun2017collaboratively}, and the category it belongs to \cite{meng2020discriminative}. Inspired by these studies, we propose the following embedding learning objective.
\begin{equation}
\small
\begin{split}
    \mathcal{J} = & \underbrace{\sum_{d\in\mathcal{D}}\sum_{w_i \in d}\ \sum_{w_j\in \mathcal{C}(w_i, h)} p(w_j|w_i)}_{\rm context} \\
    & + \underbrace{\sum_{d\in \mathcal{D}}\sum_{w\in d} p(d|w)}_{\rm document} + \underbrace{\sum_{c_i \in \mathcal{C}}\sum_{w\in \mathcal{S}_i} p(c_i|w)}_{\rm category},
\end{split}
\label{eqn:objective}
\end{equation}
where
\begin{equation}
\small
    p(z|w) = \frac{\exp(\bmu_w^T\bmv_z)}{\sum_{z'}\exp(\bmu_w^T\bmv_{z'})},\ \ \ (z \text{ can be } w_j, d, \text{ or } c_i).
\label{eqn:softmax}
\end{equation}
In this objective, $\bmu_{w_i}$ (and $\bmv_{w_j}$), $\bmv_d$, $\bmv_{c_i}$ are the embedding vectors of terms, documents, and categories, respectively. $\mathcal{C}(w_i, h)$ is the set of context terms of $w_i$ in $d$. Specifically, if $d=w_1w_2...w_L$, then $\mathcal{C}(w_i, h)=\{w_j|i-h\leq j \leq i+h,j\neq i\}$, where $h$ is the context window size.

Note that the last term in Eq. (\ref{eqn:objective}) encourages the similarity between each category $c_i$ and its representative terms $\mathcal{S}_i$. Here, we adopt an iterative process to gradually update category-representative terms. Initially, $\mathcal{S}_i$ consists of just a few in-vocabulary terms similar to $c_i$ according to the PLM. At each iteration, the size of $\mathcal{S}_i$ will increase to contain more category-discriminative terms (the selection criterion of these terms will be introduced in the next section), and we need to encourage their proximity with $c_i$ in the next iteration.

Directly optimizing the full softmax in Eq. (\ref{eqn:softmax}) is costly. Therefore, we adopt the negative sampling strategy \cite{mikolov2013distributed} for efficient approximation.

\vspace{1mm}

\noindent \textbf{Interpreting the Objective.} In topic modeling studies, pointwise mutual information (PMI) \cite{newman2010automatic} is a standard evaluation metric for topic coherence \cite{lau2014machine,roder2015exploring}. \citet{levy2014neural} prove that the Skip-Gram embedding model is implicitly factorizing the PMI matrix. Following their proof, we can show that maximizing Eq. (\ref{eqn:objective}) is implicitly doing the following factorization:
\begin{equation}
\small
    \bU_w^T [\bV_w; \bV_d; \bV_c] = [\bX_{ww}; \bX_{wd}; \bX_{wc}],
\label{eqn:matrix}
\end{equation}
where the columns of $\bU_w$, $\bV_w$, $\bV_d$, $\bV_c$ are $\bmu_{w_i}$, $\bmv_{w_j}$, $\bmv_d$, $\bmv_{c_i}$, respectively ($w_i, w_j \in \mathcal{V}_{\mathcal{D}}$, $d \in \mathcal{D}$, $c_i \in \mathcal{C}$); $\bX_{ww}$, $\bX_{wd}$, and $\bX_{wc}$ are PMI matrices.
\begin{equation}
\small
\begin{split}
\bX_{ww} &= \Bigg[\log \Big(\frac{\#_{\mathcal{D}}(w_i,w_j) \cdot \lambda_{\mathcal{D}}}{\#_{\mathcal{D}}(w_i) \cdot \#_{\mathcal{D}}(w_j) \cdot b}\Big)\Bigg]_{w_i, w_j \in \mathcal{V}_{\mathcal{D}}}, \\
\bX_{wd} &= \Bigg[\log \Big(\frac{\#_{d}(w) \cdot \lambda_{\mathcal{D}}}{\#_{\mathcal{D}}(w) \cdot \lambda_d \cdot b}\Big)\Bigg]_{w \in \mathcal{V}_{\mathcal{D}},\ d \in \mathcal{D}}, \\
\bX_{wc} &= \big[x_{w, c_i}\big]_{w \in \mathcal{V}_{\mathcal{D}},\ c_i \in \mathcal{C}}, \ \ \ \text{where}\\
&\ \ \ \ \ \ x_{w, c_i} =
\begin{cases}
\log \frac{|\mathcal{C}|}{b}, & {\rm if\ } w \in \mathcal{S}_i, \\
-\infty, & {\rm if\ } w \in \mathcal{S}_j \ (\forall j \neq i).
% , \\
% \text{missing}, & {\rm if\ } w \notin \mathcal{S}_1 \cup ... \cup \mathcal{S}_{|\mathcal{C}|}.
\end{cases}
\end{split}
\label{eqn:levy}
\end{equation}
Here, $\#_{\mathcal{D}}(w_i,w_j)$ denotes the number of co-occurrences of $w_i$ and $w_j$ in a context window in $\mathcal{D}$; $\#_{\mathcal{D}}(w)$ denotes the number of occurrences of $w$ in $\mathcal{D}$; $\lambda_{\mathcal{D}}$ is the total number of terms in $\mathcal{D}$; $\#_{d}(w)$ denotes the number of times $w$ occurs in $d$; $\lambda_d$ is the total number of terms in $d$; $b$ is the number of negative samples. (For the derivation of Eq. (\ref{eqn:matrix}), please refer to Appendix \ref{sec:app_embedding}.)

To summarize, the learned local representations $\bmu_w$ are implicitly optimized for topic coherence, where term co-occurrences are measured in context, document, and category levels.

\subsection{Ensemble Ranking}
\label{sec:ensemble}
We have obtained two sets of term embeddings that model text semantics from different angles: $\{\bme_w|w\in \mathcal{C} \cup \mathcal{V}_{\mathcal{D}}\}$ carries the PLM's knowledge, while $\{\bmu_w|w\in \mathcal{V}_{\mathcal{D}}\}$ models the input corpus as well as user-provided seeds. We now propose an ensemble ranking method to leverage information from both sides to grab more discriminative terms for each category.

Given a category $c_i$ and its current term set $\mathcal{S}_i$, we first calculate the scores of each term $w\in \mathcal{V}_{\mathcal{D}}$.
\begin{equation}
\small
\begin{split}
    {\rm score}_G(w|\mathcal{S}_i) &= \frac{1}{|\mathcal{S}_i|}\sum_{w'\in \mathcal{S}_i}\cos(\bme_w, \bme_{w'}), \\
    {\rm score}_L(w|\mathcal{S}_i) &= \frac{1}{|\mathcal{S}_i|}\sum_{w'\in \mathcal{S}_i}\cos(\bmu_w, \bmu_{w'}).
\end{split}
\label{eqn:cos}
\end{equation}
The subscript ``$G$'' here means ``general'', while ``$L$'' means ``local''. Then, we sort all terms by these two scores, respectively. Each term $w$ will hence get two rank positions ${\rm rank}_G(w)$ and ${\rm rank}_L(w)$. We propose the following ensemble score based on the reciprocal rank:
\begin{equation}
\small
    {\rm score}(w|\mathcal{S}_i) = \bigg(\frac{1}{2}\Big(\frac{1}{{\rm rank}_G(w)}\Big)^\rho + \frac{1}{2}\Big(\frac{1}{{\rm rank}_L(w)}\Big)^\rho\bigg)^{1/\rho}.
\label{eqn:holder}
\end{equation}
Here, $0 < \rho \leq 1$ is a constant. In practice, instead of ranking all terms in the vocabulary, we only check the top-$M$ results in the two ranking lists. If a term $w$ is not among the top-$M$ according to ${\rm score}_G(w)$ (resp., ${\rm score}_L(w)$), we set ${\rm rank}_G(w)=+\infty$ (resp., ${\rm rank}_L(w)=+\infty$).
In fact, when $\rho=1$, Eq. (\ref{eqn:holder}) becomes the \textit{arithmetic} mean of the two reciprocal ranks $\frac{1}{{\rm rank}_G(w)}$ and $\frac{1}{{\rm rank}_L(w)}$. This is essentially the mean reciprocal rank (MRR) commonly used in ensemble ranking, where a high position in one ranking list can largely compensate a low position in the other. In contrast, when $\rho \rightarrow 0$, Eq. (\ref{eqn:holder}) becomes the \textit{geometric} mean of the two reciprocal ranks (see Appendix \ref{sec:app_ensemble}), where two ranking lists both have the ``veto power'' (i.e., a term needs to be ranked as top-$M$ in both ranking lists to obtain a non-zero ensemble score). In experiment, we set $\rho=0.1$ and show it outperforms MRR (i.e., $\rho=1$) in our topic discovery task.

After computing the ensemble score ${\rm score}(w|\mathcal{S}_i)$ for each $w$, we update $\mathcal{S}_i$. To guarantee that each $\mathcal{S}_i$ is category-discriminative, we do not allow any term to belong to more than one category. Therefore, we gradually expand each $\mathcal{S}_i$ by turns. At the beginning, we reset $\mathcal{S}_1=...=\mathcal{S_{|\mathcal{C}|}}=\emptyset$. When it is $\mathcal{S}_i$'s turn, we add one term $\mathcal{S}_i$ according to the following criterion:
\begin{equation}
\small
    \mathcal{S}_i \leftarrow \mathcal{S}_i \cup \{\argmax_{w \in \mathcal{\mathcal{V}_{\mathcal{D}}}\backslash (\mathcal{S}_1 \cup ... \cup \mathcal{S}_{|\mathcal{C}|})} {\rm score}(w|\mathcal{S}_i)\}.
\label{eqn:update}
\end{equation}

\subsection{Overall Framework}

\newlength{\textfloatsepsave} 
\setlength{\textfloatsepsave}{\textfloatsep}
\setlength{\textfloatsep}{0pt}
\begin{algorithm}[t]
\small
\caption{\textsc{\model}}
\label{alg:overall}
\KwIn{
A text corpus $\mathcal{D}=\{d_1,...,d_{|\mathcal{D}|}\}$, a set of seeds $\mathcal{C}=\{c_1,...,c_{|\mathcal{C}|}\}$, and a PLM.
}
\KwOut{$(\mathcal{S}_1, ..., \mathcal{S}_{|\mathcal{C}|})$, where each $\mathcal{S}_i$ is a set of category-discriminative terms for $c_i$.}

Compute $\{\bme_w|w\in \mathcal{C} \cup \mathcal{V}_{\mathcal{D}}\}$ using the PLM\;
\textcolor{myblue}{// Initialize $\mathcal{S}_i$\;}
$\mathcal{S}_1,...,\mathcal{S_{|\mathcal{C}|}} \gets \emptyset$\;
\For{$n \gets 1$ to $N$}
{
    \For{$i \gets 1$ to $|\mathcal{C}|$}
    {
        $w_n \gets \argmax\limits_{w \in \mathcal{\mathcal{V}_{\mathcal{D}}}\backslash (\mathcal{S}_1 \cup ... \cup \mathcal{S}_{|\mathcal{C}|})} \cos(\bme_w, \bme_{c_i})$\;
        $\mathcal{S}_i \gets \mathcal{S}_i \cup \{w_n\}$\;
    }
}
\textcolor{myblue}{// Update $\mathcal{S}_i$ for $T$ iterations\;}
\For{$t \gets 1$ to $T$}
{
    Learn $\{\bmu_w|w\in \mathcal{V}_{\mathcal{D}}\}$ from the input corpus $\mathcal{D}$ and the up-to-date representative terms $\mathcal{S}_1, ..., \mathcal{S}_{|\mathcal{C}|}$ according to Eq. (\ref{eqn:objective})\;
    ${\rm score}_G(w|\mathcal{S}_i)$ and ${\rm score}_L(w|\mathcal{S}_i) \gets$ Eq. (\ref{eqn:cos})\;
    ${\rm score}(w|\mathcal{S}_i) \gets$ Eq. (\ref{eqn:holder})\;
    $\mathcal{S}_1,...,\mathcal{S_{|\mathcal{C}|}} \gets \emptyset$\;
    \For{$n \gets 1$ to $(t+1)N$}
    {
        \For{$i \gets 1$ to $|\mathcal{C}|$}
        {
            $\mathcal{S}_i \gets$ Eq. (\ref{eqn:update})\;
        }
    }
}
Return $(\mathcal{S}_1, ..., \mathcal{S}_{|\mathcal{C}|})$\;
\end{algorithm}
\setlength{\textfloatsep}{\textfloatsepsave}

We summarize the entire \textsc{\model} framework in Algorithm \ref{alg:overall}. To deal with \textbf{out-of-vocabulary} category names, we first utilize a PLM to find their nearest in-vocabulary terms as the initial category-discriminative term set $\mathcal{S}_i$ (Lines 1-7). After initialization, $|\mathcal{S}_i|=N$ $(\forall 1\leq i\leq |\mathcal{C}|)$. Note that for an in-vocabulary category name $c_i \in \mathcal{V}_{\mathcal{D}}$, itself will be added to the initial $\mathcal{S}_i$ as the top-1 similar in-vocabulary term.

After getting the initial $\mathcal{S}_i$, we update it by $T$ iterations (Lines 8-16). At each iteration, according to the up-to-date $\mathcal{S}_1, \mathcal{S}_2, ..., \mathcal{S}_{|\mathcal{C}|}$, we relearn embeddings $\bmu_w$, $\bmv_w$, $\bmv_d$, and $\bmv_{c_i}$ using Eq. (\ref{eqn:objective}) (Line 10). The two set of embeddings, $\{\bme_w|w\in \mathcal{C} \cup \mathcal{V}_{\mathcal{D}}\}$ (computed at Line 1) and $\{\bmu_w|w\in \mathcal{V}_{\mathcal{D}}\}$ (updated at Line 10), are then leveraged to perform ensemble ranking (Lines 11-12). Based on the ensemble score ${\rm score}(w|\mathcal{S}_i)$, we update $\mathcal{S}_i$ using Eq. (\ref{eqn:update}) (Lines 13-16). After the $t$-th iteration, $|\mathcal{S}_i|=(t+1)N$ $(\forall 1\leq i\leq |\mathcal{C}|)$.

\vspace{1mm}

\noindent \textbf{Complexity Analysis.} The time complexity of using the PLM is $\mathcal{O}((|\mathcal{C}| + |\mathcal{V}_{\mathcal{D}}|)\alpha_{\rm PLM})$, where $\alpha_{\rm PLM}$ is the complexity of encoding one term via the PLM.
The total complexity of local embedding is $\mathcal{O}(T\lambda_{\mathcal{D}} (h+|\mathcal{C}|) b)$ because in each iteration $1\le t \le T$, every $w \in \mathcal{D}$ interacts with every other term in the context window of size $h$, its belonging document, and each category $c_i \in \mathcal{C}$, and each update involves $b$ negative samples.
The total complexity of ensemble ranking is $\mathcal{O}(T|\mathcal{V}_{\mathcal{D}}||\mathcal{C}||\mathcal{S}_i|)$ as in each iteration $1\le t \le T$, we compute scores between each $w \in \mathcal{V}_{\mathcal{D}}$ and each $w' \in \mathcal{S}_i\  (\forall c_i \in \mathcal{C})$.

\section{Experiments}
\subsection{Experimental Setup}
\noindent \textbf{Datasets.} We conduct experiments on three public datasets from different domains:
(1) \underline{\textbf{SciDocs}} \cite{cohan2020specter}\footnote{\scriptsize \url{https://github.com/allenai/scidocs}} is a large collection of scientific papers supporting diverse evaluation tasks. For the MeSH classification task \cite{coletti2001medical}, about 23K medical papers are collected, each of which is assigned to one of the 11 common disease categories derived from the MeSH vocabulary. We use the title and abstract of each paper as documents and the 11 category names as seeds. 
(2) \underline{\textbf{Amazon}} \cite{mcauley2013hidden}\footnote{\scriptsize \url{http://jmcauley.ucsd.edu/data/amazon/index.html}} contains product reviews from May 1996 to July 2014. Each Amazon review belongs to one or more product categories. We use the subset sampled by \citet{zhang2020minimally,zhang2022motifclass}, which contains 10 categories and 100K reviews.
(3) \underline{\textbf{Twitter}} \cite{zhang2017react}\footnote{\scriptsize \url{https://github.com/franticnerd/geoburst}} is a crawl of geo-tagged tweets in New York City from August 2014 to November 2014. The dataset collectors link these tweets with Foursquare’s POI database and assign them to 9 POI categories. We take these category names as input seeds.

Seeds used in the three datasets are shown in Table \ref{tab:intro}. Dataset statistics are summarized in Table \ref{tab:data}. For all three datasets, we use AutoPhrase \cite{shang2018automated}\footnote{\scriptsize \url{https://github.com/shangjingbo1226/AutoPhrase}} to perform phrase chunking in the corpus, and we remove words and phrases occurring less than 3 times.

Previous studies \cite{jagarlamudi2012incorporating,meng2020discriminative} have tried some other datasets (e.g., RCV1, 20 Newsgroups, NYT, and Yelp). However, the category names they use in these datasets are all picked from \textbf{in-vocabulary} terms. Therefore, we do not consider these datasets for evaluation in our task settings.

Following \cite{sia2020tired}, we adopt a 60-40 train-test split for all three datasets. The training set is used as the input corpus $\mathcal{D}$, and the testing set is used for calculating topic coherence metrics (see evaluation metrics for details).

\begin{table}[t]
\centering
\caption{Dataset Statistics.}
\scalebox{0.73}{
\begin{tabular}{>{\centering\arraybackslash}m{4.1cm}|>{\centering\arraybackslash}m{1.5cm}|>{\centering\arraybackslash}p{1.5cm}|>{\centering\arraybackslash}m{1.5cm}}
\hline
\textbf{Dataset}          & \textbf{SciDocs}                 & \textbf{Amazon}   & \textbf{Twitter}      \\ \hline
\#Documents      & 23,473 & 100,000 & 135,529      \\ \hline
\#In-vocabulary Terms (After Phrase Chunking) & 55,897 & 56,942 & 17,577            \\ \hline
Avg Doc Length   & 239.8 & 119.0 & 6.7   \\ \hline
\#Seeds     & 11 & 10 & 9         \\ \hline
\#Out-of-vocabulary Seeds (After Phrase Chunking) & 5 & 6 & 7            \\ \hline
\end{tabular}
}
\vspace{-0.5em}
\label{tab:data}
\end{table}

\begin{table*}[]
\centering
\caption{NPMI, LCP, MACC, and Diversity of compared algorithms on three datasets. NPMI and LCP measure topic coherence; MACC measures term accuracy; Diversity (abbreviated to Div.) measures topic diversity. \textbf{Bold}: the highest score. \underline{Underline}: the second highest score. $^{*}$: significantly worse than \textsc{\model} (p-value $< 0.05$). $^{**}$: significantly worse than \textsc{\model} (p-value $< 0.01$).}
\scalebox{0.69}{
\begin{tabular}{c|cccc|cccc|cccc}
\hline
\multirow{2}{*}{Methods} & \multicolumn{4}{c|}{SciDocs}                 & \multicolumn{4}{c|}{Amazon}                  & \multicolumn{4}{c}{Twitter}                  \\ \cline{2-13} 
                         & NPMI             & LCP           & MACC   & Div.   & NPMI             & LCP           & MACC  & Div.    & NPMI             & LCP           & MACC  & Div.    \\ \hline
SeededLDA                & 0.111$^{**}$            & \textbf{-1.232}           &   0.156$^{**}$   & 0.451$^{**}$     & 0.140$^{**}$           & \underline{-1.505}           &    0.147$^{**}$  & 0.393$^{**}$    &   0.026$^{**}$             &  -4.508$^{**}$      &    0.195$^{**}$    & 0.696$^{**}$   \\
Anchored CorEx           & 0.213$^{**}$           & -2.180$^{**}$           &   0.264$^{**}$  & \textbf{1.000}      & 0.268$^{**}$           & -1.963$^{*}$          &  0.333$^{**}$    & \textbf{1.000}    &   0.180$^{**}$            &  -4.384$^{**}$      &   0.233$^{**}$    & \textbf{1.000}   \\
Labeled ETM              & 0.669$^{*}$          & -1.549$^{**}$          &  0.458$^{**}$    & 0.961$^{*}$     & 0.616$^{**}$        & -2.103$^{**}$         &    0.585$^{**}$   & \textbf{1.000}    &   0.610$^{*}$        &   -2.197$^{**}$        &    0.268$^{**}$   & 0.989    \\
CatE                     & \underline{0.691}$^{*}$          & -1.451$^{**}$        &    0.633$^{**}$    &  \textbf{1.000}  & 0.633$^{**}$          & -1.688$^{**}$          &  0.856$^{*}$    & \textbf{1.000}     &     \textbf{0.711}   &    \textbf{-1.653}       &  0.483$^{**}$    & \textbf{1.000}     \\ \hline
BERT                     & 0.626$^{**}$          & -1.682$^{**}$          &  0.740$^{**}$   & 0.891$^{**}$      & 0.588$^{**}$          & -2.186$^{**}$          &    0.832$^{**}$   & \textbf{1.000}    & 0.626$^{**}$          & -2.088$^{**}$          &    \underline{0.627}   & 0.944$^{**}$    \\
BioBERT                  & 0.618$^{**}$          & -1.704$^{**}$          &  \textbf{0.938}   & 0.982$^{**}$      & \textbf{--}               & \textbf{--}              & \textbf{--}         & \textbf{--}               & \textbf{--}            & \textbf{--}   & \textbf{--} & \textbf{--}       \\ \hline
\textsc{\model}-NoIter &  0.681$^{**}$  &  -1.535$^{**}$  &  0.887  &  \textbf{1.000}  & \underline{0.643}$^{**}$ & -1.971$^{**}$ &  \underline{0.892}  &  \textbf{1.000}  & 0.636 & -2.008$^{**}$ &  0.618 &  \textbf{1.000} \\
\textsc{\model}              & \textbf{0.717} & \underline{-1.267} & \underline{0.909} & \textbf{1.000} & \textbf{0.684} & \textbf{-1.391} & \textbf{0.904} & \textbf{1.000} & \underline{0.640} & \underline{-1.813} & \textbf{0.633} & \textbf{1.000} \\ \hline
\end{tabular}
}
\vspace{-0.5em}
\label{tab:performance}
\end{table*}

\vspace{1mm}

\noindent \textbf{Compared Methods.} We compare our \textsc{\model} framework with the following methods, including seed-guided topic modeling methods, seed-guided embedding learning methods, and PLMs.
(1) \underline{\textbf{SeededLDA}} \cite{jagarlamudi2012incorporating}\footnote{\scriptsize \url{https://github.com/vi3k6i5/GuidedLDA}} is a seed-guided topic modeling method. It improves LDA by biasing topics to produce input seeds and by biasing documents to select topics relevant to the seeds they contain.
(2) \underline{\textbf{Anchored CorEx}} \cite{gallagher2017anchored}\footnote{\scriptsize \url{https://github.com/gregversteeg/corex_topic}} is a seed-guided topic modeling method. It incorporates user-provided seeds by balancing between compressing the input corpus and preserving seed-related information.
(3) \underline{\textbf{Labeled ETM}} \cite{dieng2020topic}\footnote{\scriptsize \url{https://github.com/adjidieng/ETM}} is an embedding-based topic modeling method. It incorporates distributed representation of each term. Following \cite{meng2020discriminative}, we retrieve representative terms according to their embedding similarity with the category name.
(4) \underline{\textbf{CatE}} \cite{meng2020discriminative}\footnote{\scriptsize \url{https://github.com/yumeng5/CatE}} is a seed-guided embedding learning method for discriminative topic discovery. It takes category names as input and jointly learns term embedding and specificity from the input corpus. Category-discriminative terms are then selected based on both embedding similarity with the category and specificity.
(5) \underline{\textbf{BERT}} \cite{devlin2019bert}\footnote{\scriptsize \url{https://huggingface.co/bert-base-uncased}} is a PLM. Following Lines 1-7 in Algorithm \ref{alg:overall}, we use BERT to encode each input category name and each term to a vector, and then perform similarity search to directly find all representative terms. 
(6) \underline{\textbf{BioBERT}} \cite{lee2020biobert}\footnote{\scriptsize \url{https://huggingface.co/dmis-lab/biobert-v1.1}} is a PLM. It is used in the same way as BERT. Since BioBERT is specifically trained for biomedical text mining tasks, we report its performance on the SciDocs dataset only.
(7) \underline{\textbf{\textsc{\model}-NoIter}} is a variant of our \textsc{\model} framework. In Algorithm \ref{alg:overall}, after initialization (Lines 1-7), it executes Lines 9-16 only once (i.e., $T=1$) to find all representative terms.

Here, all seed-guided topic modeling and embedding baselines (i.e., SeededLDA, Anchored CorEx, CatE, and Labeled ETM) can only take \textbf{in-vocabulary} seeds as input. For a fair comparison, we run Lines 1-7 in Algorithm \ref{alg:overall} to get the initial representative in-vocabulary terms for each category, and input these terms as seeds into the baselines. In other words, all compared methods use BERT/BioBERT to initialize their term sets.

\vspace{1mm}

\noindent \textbf{Evaluation Metrics.} We evaluate topic discovery results from three different angles: topic coherence, term accuracy, and topic diversity.

\vspace{1mm}

\noindent (1) \underline{\textbf{NPMI}} \cite{lau2014machine} is a standard metric in topic modeling to measure \underline{\textit{topic coherence}}. Within each topic, it calculates the normalized pointwise mutual information for each pair of terms in $\mathcal{S}_i$.
\begin{equation}
\small
    {\rm NPMI} = \frac{1}{|\mathcal{C}|}\sum_{i=1}^{|\mathcal{C}|} \frac{1}{\binom{|\mathcal{S}_i|}{2}}\sum_{\substack{w_j, w_k \in \mathcal{S}_i \\ j < k}} \frac{\log\frac{P(w_j, w_k)}{P(w_j)P(w_k)}}{-\log P(w_j, w_k)},
    \label{eqn:npmi}
\end{equation}
where $P(w_j, w_k)$ is the probability that $w_j$ and $w_k$ co-occur in a document; $P(w_j)$ is the marginal probability of $w_j$.\footnote{When calculating Eqs. (\ref{eqn:npmi}) and (\ref{eqn:lcp}), to avoid $\log 0$, we use $P(w_j, w_k)+\epsilon$ and $P(w)+\epsilon$ to replace $P(w_j, w_k)$ and $P(w)$, respectively, where $\epsilon = 1/|\mathcal{D}|$.}

\vspace{1mm}
    
\noindent (2) \underline{\textbf{LCP}} \cite{mimno2011optimizing} is another standard metric to measure \underline{\textit{topic coherence}}. It calculates the pairwise log conditional probability of top-ranked terms.
\begin{equation}
\small
    {\rm LCP} = \frac{1}{|\mathcal{C}|}\sum_{i=1}^{|\mathcal{C}|} \frac{1}{\binom{|\mathcal{S}_i|}{2}}\sum_{\substack{w_j, w_k \in \mathcal{S}_i \\ j < k}} \log\frac{P(w_j, w_k)}{P(w_j)}.
    \label{eqn:lcp}
\end{equation}
Note that \textbf{PMI} \cite{newman2010automatic} is also a standard metric for topic coherence. We do observe that \textsc{\model} outperforms baselines in terms of PMI in most cases. However, since our local embedding step is implicitly optimizing a PMI-like objective, we no longer use it as our evaluation metric.

\vspace{1mm}

\noindent (3) \underline{\textbf{MACC}} \cite{meng2020discriminative} measures \underline{\textit{term ac-}} \underline{\textit{curacy}}. It is defined as the proportion of retrieved terms that actually belong to the corresponding category according to the category name.
\begin{equation}
\small
    {\rm MACC} = \frac{1}{|\mathcal{C}|}\sum_{i=1}^{|\mathcal{C}|}\frac{1}{|\mathcal{S}_i|}\sum_{w_j \in \mathcal{S}_i} {\bf 1}(w_j \in c_i),
    \label{eqn:macc}
\end{equation}
where ${\bf 1}(w_j \in c_i)$ is the indicator function of whether $w_j$ is relevant to category $c_i$. MACC requires human evaluation, so we invite five annotators to perform independent annotation. The reported MACC score is the average MACC of the five annotators. A high inter-annotator agreement is observed, with Fleiss' kappa \cite{fleiss1971measuring} being 0.856, 0.844, and 0.771 on SciDocs, Amazon, and Twitter, respectively.

\vspace{1mm}

\noindent (4) \underline{\textbf{Diversity}} \cite{dieng2020topic} measures the \underline{\textit{mutual exclusivity}} of discovered topics. It is the percentage of unique terms in all topics, which corresponds to our task requirement that each retrieved term is discriminatively close to one category and far from the others.
\begin{equation}
\small
    {\rm Diversity} = \frac{|\bigcup_{i=1}^{|\mathcal{C}|}\mathcal{S}_i|}{\sum_{i=1}^{|\mathcal{C}|}|\mathcal{S}_i|}.
    \label{eqn:dist}
\end{equation}

\vspace{1mm}

\noindent \textbf{Experiment Settings.} We use BioBERT as the PLM on SciDocs, and BERT-base-uncased as the PLM on Amazon and Twitter. The embedding dimension of $\bmu_w$ is 768 (the same as $\bme_w$); the number of negative samples $b=5$. In ensemble ranking, the length of the general/local ranking list $M=100$; the hyperparameter $\rho$ in Eq. (\ref{eqn:holder}) is set as $0.1$; the number of iterations $T=4$; after each iteration, we increase the size of $\mathcal{S}_i$ by $N=3$. We use the top-10 ranked terms in each topic for final evaluation (i.e., $|\mathcal{S}_i|=10$ in Eqs. (\ref{eqn:npmi})-(\ref{eqn:dist})). Experiments are run on Intel Xeon E5-2680 v2 @ 2.80GHz and one NVIDIA GeForce GTX 1080.

\subsection{Performance Comparison}
Table \ref{tab:performance} shows the performance of all methods. We run each experiment 3 times with the average score reported. 
To show statistical significance, we conduct a two-tailed unpaired t-test to compare \textsc{\model} and each baseline. (The performance of BERT and BioBERT is deterministic according to our usage. When comparing \textsc{\model} with them, we conduct a two-tailed Z-test instead.) The significance level is also marked in Table \ref{tab:performance}.

We have the following observations from Table \ref{tab:performance}. (1) Our \textsc{\model} model performs consistently well. In fact, it achieves the highest score in 8 columns and the second highest in the remaining 4 columns. (2) Classical seed-guided topic modeling baselines (i.e., SeededLDA and Anchored CorEx) perform not well in respect of NPMI (topic coherence) and MACC (term accuracy). Embedding-based topic discovery approaches (i.e., Labeled ETM and CatE) make some progress, but they still significantly underperform the PLM-empowered \textsc{\model} model on SciDocs and Amazon. (3) \textsc{\model} consistently performs better than \textsc{\model}-NoIter on all three datasets, indicating the positive contribution of the proposed iterative process. (4) \textsc{\model} guarantees the mutual exclusivity of $\mathcal{S}_1,...,\mathcal{S}_{|\mathcal{C}|}$.  In comparison, SeededLDA, Labeled ETM, and BERT cannot guarantee such mutual exclusivity.

\vspace{1mm}

\begin{figure}
\centering
\subfigure[SciDocs]{
\includegraphics[width=0.23\textwidth]{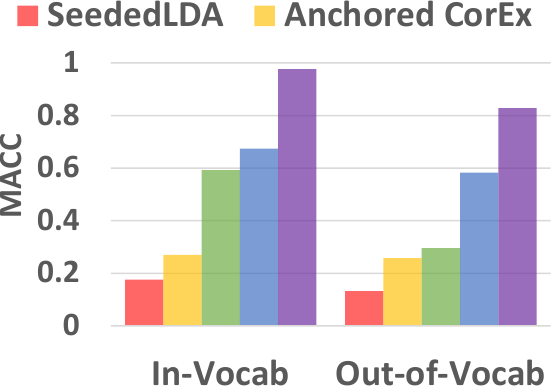}}
\hspace{-0.5mm}
\subfigure[Amazon]{
\includegraphics[width=0.23\textwidth]{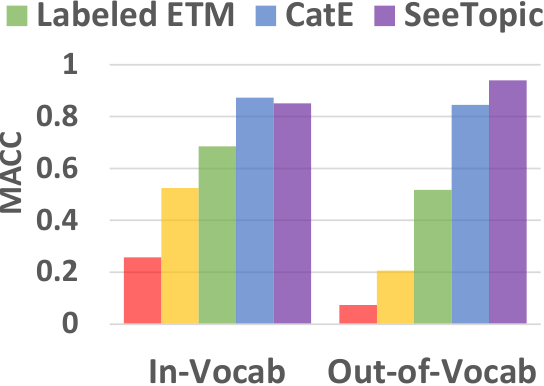}} \\
\subfigure[Twitter]{
\includegraphics[width=0.23\textwidth]{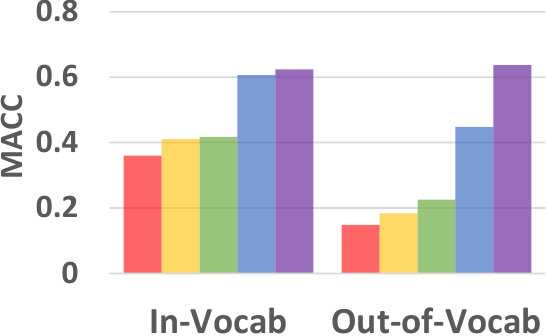}}
\vspace{-0.5em}
\caption{MACC of seed-guided topic discovery methods on in-vocabulary categories and out-of-vocabulary categories.} 
\vspace{-0.5em}
\label{fig:oov}
\end{figure}

\noindent \textbf{In-vocabulary vs. Out-of-vocabulary.} Figure \ref{fig:oov} compares the MACC scores of different seed-guided topic discovery methods on in-vocabulary categories and out-of-vocabulary categories. We find that the performance improvement of \textsc{\model} upon baselines on out-of-vocabulary categories is larger than that on in-vocabulary ones. For example, on Amazon, \textsc{\model} underperforms CatE on in-vocabulary categories but outperforms CatE on out-of-vocabulary ones; on Twitter, the gap between \textsc{\model} and baselines becomes much more evident on out-of-vocabulary categories. Note that all baselines in Figure \ref{fig:oov} do not utilize the power of PLMs, so this observation validates our claim that PLMs are helpful in tackling out-of-vocabulary seeds.

\subsection{Parameter Study}
\begin{figure}
\centering
\subfigure[Effect of $\rho$ on NPMI]{
\includegraphics[width=0.23\textwidth]{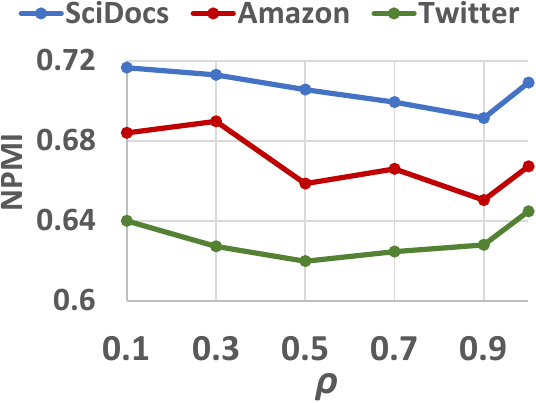}}
\hspace{-0.5mm}
\subfigure[Effect of $\rho$ on LCP]{
\includegraphics[width=0.23\textwidth]{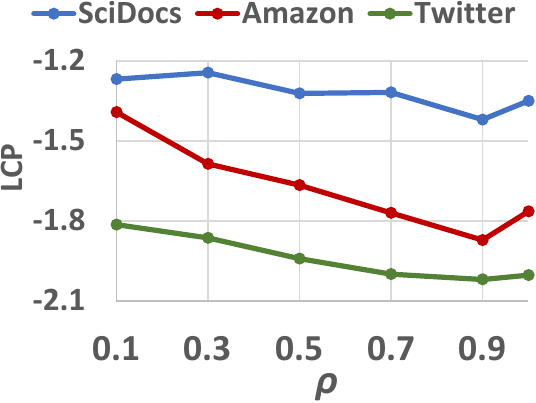}} \\
\subfigure[Effect of $T$ on NPMI]{
\includegraphics[width=0.23\textwidth]{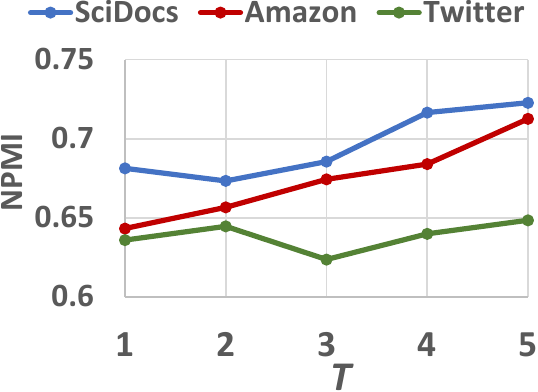}}
\hspace{-0.5mm}
\subfigure[Effect of $T$ on LCP]{
\includegraphics[width=0.23\textwidth]{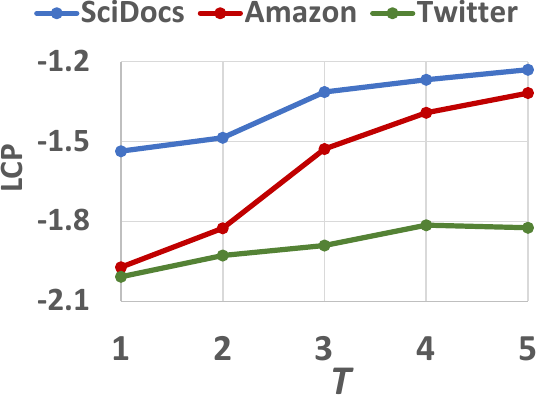}}
\vspace{-0.5em}
\caption{Parameter study of \textsc{\model} measured by topic coherence.} 
\vspace{-0.8em}
\label{fig:para}
\end{figure}

We study the effect of two important hyperparameters: $\rho$ (the hyperparameter in ensemble ranking) and $T$ (the number of iterations). We vary the value of $\rho$ in $\{0.1, 0.3, 0.5, 0.7, 0.9, 1\}$ (\textsc{\model} uses $\rho=0.1$ by default) and the value of $T$ in $\{1, 2, 3, 4, 5\}$ (\textsc{\model} uses $T=4$ by default, and \textsc{\model}-NoIter is the case when $T=1$). Figure \ref{fig:para} shows the change of model performance measured by NPMI and LCP.

\begin{table*}[t]
\centering
\caption{Top-5 representative terms retrieved by different algorithms for three out-of-vocabulary categories from SciDocs, Amazon, and Twitter. \cmark: at least 3 of the 5 annotators judge the term as relevant to the seed. \xmark: at most 2 of the 5 annotators judge the term as relevant to the seed.}
\vspace{-0.5em}
\scalebox{0.73}{
\begin{tabular}{c|c}
\hline
\textbf{Method}         & \textbf{Top-5 Representative Terms}                                                                                                                                           \\ \hline \hline
\multicolumn{2}{c}{\cellcolor{black!20} Dataset: SciDocs, \ \ \ Category Name: hepatitis a/b/c/e} \\ \hline
SeededLDA      & patients (\xmark), treatment (\xmark), placebo (\xmark), study (\xmark), group (\xmark)                                                                        \\ \hline
Anchored CorEx & expression (\xmark), gene (\xmark), cells (\xmark), genes (\xmark), genetic (\xmark)                                                                           \\ \hline
Labeled ETM    & \begin{tabular}[c]{@{}c@{}} hepatitis b virus hbv dna (\cmark), serum hbv dna (\cmark), serum alanine aminotransferase (\xmark), \\ alanine aminotransferase alt (\xmark), below detection limit (\xmark)   \end{tabular}                                     \\ \hline
CatE           & \begin{tabular}[c]{@{}c@{}} chronic hepatitis b virus hbv infection (\cmark),	hepatitis b e antigen hbeag (\cmark),	hepatitis b virus hbv dna (\cmark), \\	normal alanine aminotransferase (\xmark),	hbeag-negative chronic hepatitis b (\cmark)   \end{tabular}                                                                      \\ \hline
BioBERT        & \begin{tabular}[c]{@{}c@{}} hepatitis b virus hbv dna (\cmark),	chronic hepatitis b virus hbv infection (\cmark),	hepatitis b e antigen hbeag (\cmark), \\	hepatitis b virus hbv infection (\cmark),	chronic hepatitis c virus hcv (\cmark) \end{tabular}  \\ \hline
\textsc{\model}-NoIter        & \begin{tabular}[c]{@{}c@{}} hepatitis b virus hbv dna (\cmark),	hepatitis b e antigen hbeag (\cmark),	chronic hepatitis b virus hbv infection (\cmark), \\	hepatitis b surface antigen hbsag (\cmark),	hbeag-negative chronic hepatitis b (\cmark) \end{tabular} \\ \hline
\textsc{\model}        & \begin{tabular}[c]{@{}c@{}} chronic hepatitis b virus hbv infection (\cmark),	hbeag-negative chronic hepatitis b (\cmark), hepatitis c virus hcv-infected (\cmark), \\	hepatitis b virus hbv dna (\cmark),	chronic hepatitis c virus hcv (\cmark) \end{tabular} \\ \hline \hline 
\multicolumn{2}{c}{\cellcolor{black!20} Dataset: Amazon, \ \ \ Category Name: sports and outdoors} \\ \hline
SeededLDA      & use (\xmark), good (\xmark), one (\xmark), product (\xmark), like (\xmark)                                                \\ \hline
Anchored CorEx & sports (\cmark), use (\xmark), size (\xmark), wear (\xmark), fit (\cmark)                                                \\ \hline
Labeled ETM    & cars and tracks (\cmark), tracks and cars (\cmark), search options (\xmark), championships (\xmark), cool bosses (\xmark) \\ \hline
CatE           & outdoorsmen (\cmark),	outdoor activities (\cmark),	cars and tracks (\cmark),	foot support (\cmark),	offers plenty (\xmark)                        \\ \hline
BERT           & cars and tracks (\cmark),	outdoor activities (\cmark),	outdoorsmen (\cmark),	sports (\cmark),	sporting events (\cmark)       \\ \hline
\textsc{\model}-NoIter        & outdoorsmen (\cmark),	outdoor activities (\cmark),	cars and tracks (\cmark),	indoor soccer (\cmark), 	bike riding (\cmark)           \\ \hline
\textsc{\model}        & canoeing (\cmark), 	picnics (\cmark),	bike rides (\cmark),	bike riding (\cmark),	rafting (\cmark)                         \\ \hline 
\hline
\multicolumn{2}{c}{\cellcolor{black!20} Dataset: Twitter, \ \ \ Category Name: travel and transport} \\ \hline
SeededLDA      & nyc (\xmark),	new york (\xmark),	line (\cmark),	high (\xmark),	time square (\cmark)                                               \\ \hline
Anchored CorEx & new york (\xmark),	post photo (\cmark),	new (\xmark),	day (\xmark),	today (\xmark)                                                \\ \hline
Labeled ETM    & tourism (\cmark),	theview (\cmark),	file (\xmark),	morning view (\cmark),	gma (\xmark) \\ \hline
CatE           & maritime (\cmark),	tourism (\cmark),	natural history (\xmark),	scenery (\cmark),	elevate  (\xmark)    \\ \hline
BERT           & maritime (\cmark),	tourism (\cmark),	natural history (\xmark),	olive oil (\xmark),	baggage claim (\cmark)     \\ \hline
\textsc{\model}-NoIter        & maritime (\cmark),	tourism	(\cmark), natural history (\xmark),	scenery (\cmark),	navy (\xmark)          \\ \hline
\textsc{\model}        & wildlife (\cmark),	scenery (\cmark),	maritime (\cmark),	highlinepark (\xmark),	aquarium  (\cmark)                      \\ \hline
\end{tabular}
}
\vspace{-0.5em}
\label{tab:case}
\end{table*}

According to Figures \ref{fig:para}(a) and \ref{fig:para}(b), in most cases, the performance of \textsc{\model} deteriorates as $\rho$ increases from 0.1 to 0.9. Thus, setting $\rho=0.1$ always leads to competitive NPMI and LCP scores on the three datasets. Although $\rho = 1$ is better than $\rho=0.9$, its performance is still suboptimal in comparison with $\rho=0.1$. This finding indicates that replacing the mean reciprocal rank (i.e., $\rho=1$) with our proposed Eq. (\ref{eqn:holder}) is reasonable. According to Figures \ref{fig:para}(c) and \ref{fig:para}(d), \textsc{\model} usually performs better when there are more iterations. On SciDocs and Twitter, the scores start to converge after $T=4$. Besides, more iterations will result in longer running time. Overall, we believe setting $T=4$ strikes a good balance.

\subsection{Case Study}
Finally, we show the terms retrieved by different methods as a case study. From each of the three datasets, we select an out-of-vocabulary category and show its topic discovery results in Table \ref{tab:case}. We mark a retrieved term as correct (\cmark) if at least 3 of the 5 annotators judge the term as relevant to the seed. Otherwise, we mark the term as incorrect (\xmark).

For the category ``\textit{hepatitis a/b/c/e}'' from SciDocs, SeededLDA and Anchored CorEx can only find very general medical terms, which are relevant to all seeds in SciDocs and thus inaccurate; Labeled ETM and CatE find terms about ``\textit{alanine aminotransferase}'', whose elevation suggest not only hepatitis but also other diseases like diabetes and heart failure, thus not discriminative either; BioBERT and \textsc{\model}, with the power of a PLM, can accurately pick terms relevant to ``\textit{hepatitis b}'' and ``\textit{hepatitis c}''.
For the category ``\textit{sports and outdoors}'' from Amazon, SeededLDA and Anchored CorEx again find very general terms, most of which are not category-discriminative; Labeled ETM and CatE are able to pick more specific terms such as ``\textit{cars and tracks}'', but they still make mistakes; BERT, as a PLM, can accurately find terms that have lexical overlap with the category name (e.g., ``\textit{outdoorsmen}'', ``\textit{sporting events}''), meanwhile such terms are less diverse; \textsc{\model}-NoIter starts to discover more concrete terms than BERT (e.g., ``\textit{indoor soccer}'', ``\textit{bike riding}'') by leveraging local text semantics; the full \textsc{\model} model, with an iterative updating process, can find more specific and informative terms (e.g., ``\textit{canoeing}'', ``\textit{picnics}'', and ``\textit{rafting}'').
For the category ``\textit{travel and transport}'' from Twitter, both BERT and CatE make mistakes by including the term ``\textit{natural history}''; \textsc{\model}-NoIter, without an iterative update process, also includes this error; the full \textsc{\model} model finally excludes this error and achieves the highest accuracy in the retrieved top-5 terms among all compared methods.

\section{Related Work}
\noindent \textbf{Seed-Guided Topic Discovery.}
Seed-guided topic models aim to leverage user-provided seeds to discover underlying topics according to users' interests. Early studies take LDA \cite{blei2003latent} as the backbone and incorporate seeds into model learning. For example, \citet{andrzejewski2009incorporating} consider must-link and cannot-link constraints among seeds as priors. SeededLDA \cite{jagarlamudi2012incorporating} encourages topics to contain more seeds and encourages documents to select topics relevant to the seeds they contain. Anchored CorEx \cite{gallagher2017anchored} extracts maximally informative topics by jointly compressing the corpus and preserving seed relevant information. Recent studies start to utilize embedding techniques to learn better word semantics. For example, CatE \cite{meng2020discriminative} explicitly encourages distinction among retrieved topics via category-name guided embedding learning. However, all these models require the provided seeds to be in-vocabulary, mainly because they focus on the input corpus only and are not equipped with general knowledge of PLMs. 

\vspace{1mm}

\noindent \textbf{Embedding-Based Topic Discovery.}
A number of studies extend LDA to involve word embedding. The common strategy is to adapt distributions in LDA to generate real-valued data (e.g., Gaussian LDA \cite{das2015gaussian}, LFTM \cite{nguyen2015improving}, Spherical HDP \cite{batmanghelich2016nonparametric}, and CGTM \cite{xun2017correlated}). Some other studies think out of the LDA backbone. For example, TWE \cite{liu2015topical} uses topic structures to jointly learn topic embeddings and improve word embeddings. CLM \cite{xun2017collaboratively} collaboratively improves topic modeling and word embedding by coordinating global and local contexts. ETM \cite{dieng2020topic} models word-topic correlations via word embeddings to improve the expressiveness of topic models. More recently, \citet{sia2020tired} show that directly clustering word embeddings (e.g., word2vec or BERT) also generates good topics; \citet{thompson2020topic} further find that BERT and GPT-2 discover high-quality topics, but RoBERTa does not. These models are unsupervised and hard to be applied to seed-guided settings. In contrast, our \textsc{\model} framework joint leverages PLMs, word embeddings, and seed information.

\section{Conclusions and Future Work}
In this paper, we study seed-guided topic discovery in the presence of out-of-vocabulary seeds. To understand and make use of in-vocabulary components in each seed, we utilize the tokenization and contextualization power of PLMs. We propose a seed-guided embedding learning framework inspired by the goal of maximizing PMI in topic modeling, and an iterative ensemble ranking process to jointly leverage general knowledge of the PLM and local signals learned from the input corpus. Experimental results show that \textsc{\model} outperforms seed-guided topic discovery baselines and PLMs in terms of topic coherence, term accuracy, and topic diversity. A parameter study and a case study further validate some design choices in \textsc{\model}.

% There are two possible future directions in our view. 
In the future, it would be interesting to extend \textsc{\model} to seed-guided hierarchical topic discovery, where parent and child information in the input category hierarchy may help infer the meaning of out-of-vocabulary nodes. 
% Second, one may extend \textsc{\model} to not only focus on user-provided seeds but also discover other latent topics that are ``siblings'' of the provided categories, and it is worth exploring whether and how PLMs can play a role in this setting.

\section*{Acknowledgments}
We thank anonymous reviewers for their valuable and insightful feedback. Research was supported in part by US DARPA KAIROS Program No. FA8750-19-2-1004, SocialSim Program No. W911NF-17-C-0099, and INCAS Program No. HR001121C0165, National Science Foundation IIS-19-56151, IIS-17-41317, and IIS 17-04532, and the Molecule Maker Lab Institute: An AI Research Institutes program supported by NSF under Award No. 2019897, and the Institute for Geospatial Understanding through an Integrative Discovery Environment (I-GUIDE) by NSF under Award No. 2118329. Any opinions, findings, and conclusions or recommendations expressed herein are those of the authors and do not necessarily represent the views, either expressed or implied, of DARPA or the U.S. Government.
\end{spacing}

\bibliography{naacl}
\bibliographystyle{acl_natbib}

\appendix
\section{The Embedding Learning Objective}
\label{sec:app_embedding}
In Section \ref{sec:local}, we propose the following embedding learning objective:
\begin{equation}
\small
\begin{split}
    \mathcal{J} = & \underbrace{\sum_{d\in\mathcal{D}}\sum_{w_i \in d}\ \sum_{w_j\in \mathcal{C}(w_i, h)} \frac{\exp(\bmu_{w_i}^T\bmv_{w_j})}{\sum_{w'\in \mathcal{V}_{\mathcal{D}}}\exp(\bmu_{w_i}^T\bmv_{w'})}}_{\mathcal{J}_{\rm context}} + \\
    & \underbrace{\sum_{d\in \mathcal{D}}\sum_{w\in d} \frac{\exp(\bmu_w^T\bmv_d)}{\sum_{d' \in \mathcal{D}}\exp(\bmu_w^T\bmv_{d'})}}_{\mathcal{J}_{\rm document}} + \\
    & \underbrace{\sum_{c_i \in \mathcal{C}}\sum_{w\in \mathcal{S}_i} \frac{\exp(\bmu_w^T\bmv_{c_i})}{\sum_{c' \in \mathcal{C}}\exp(\bmu_w^T\bmv_{c'})}}_{\mathcal{J}_{\rm category}}.
\end{split}
\label{eqn:obj2}
\end{equation}
Now we prove that maximizing $\mathcal{J}$ is implicitly performing the factorization in Eq. (\ref{eqn:matrix}).

\citet{levy2014neural} have proved that maximizing $\mathcal{J}_{\rm context}$ is implicitly doing the following factorization.
\begin{equation}
\small
\begin{split}
\bmu_{w_i}^T\bmv_{w_j} &= \log\Big(\frac{\#_{\mathcal{D}}(w_i,w_j) \cdot \lambda_{\mathcal{D}}}{\#_{\mathcal{D}}(w_i) \cdot \#_{\mathcal{D}}(w_j) \cdot b}\Big), \\ \text{i.e., \ \ } \bU_w^T\bV_w &= \bX_{ww}.
\end{split}
\label{eqn:levy0}
\end{equation}
We follow their approach to consider the other two terms $\mathcal{J}_{\rm document}$ and $\mathcal{J}_{\rm category}$ in Eq. (\ref{eqn:obj2}). Using the negative sampling strategy to rewrite $\mathcal{J}_{\rm document}$, we get
\begin{equation}
\small
\sum_{w\in \mathcal{V}_{\mathcal{D}}}\sum_{d\in \mathcal{D}} \#_d(w) \Big(\log \sigma(\bmu_w^T\bmv_d)+b\mathbb{E}_{d'}\big[\log \sigma(-\bmu_w^T\bmv_{d'})\big]\Big),
\end{equation}
where $\sigma(\cdot)$ is the sigmoid function.
Following \cite{levy2014neural,qiu2018network}, we assume the negative sampling distribution $\propto \lambda_d$.\footnote{In practice, the negative sampling distribution $\propto \lambda_d^{3/4}$, but related studies \cite{levy2014neural,qiu2018network} usually assume a linear relationship in their derivation.} Then, the objective becomes
\begin{equation}
\small
\begin{split}
& \sum_{w\in \mathcal{V}_{\mathcal{D}}} \sum_{d\in \mathcal{D}} \#_d(w) \log \sigma(\bmu_w^T\bmv_d)\ + \\
& \sum_{w\in \mathcal{V}_{\mathcal{D}}} \#_{\mathcal{D}}(w) \sum_{d'\in \mathcal{D}}\frac{b\cdot\lambda_{d'}}{\lambda_{\mathcal{D}}}\log \sigma(-\bmu_w^T\bmv_{d'}).
\end{split}
\end{equation}
For a specific term-document pair $(w, d)$, we consider its effect in the objective:
\begin{equation}
\small
\mathcal{J}_{w,d} = \#_d(w)\log \sigma(\bmu_w^T\bmv_d)+\#_{\mathcal{D}}(w)\frac{b\cdot\lambda_d}{\lambda_{\mathcal{D}}}\log \sigma(-\bmu_w^T\bmv_d).
\end{equation}
Let $x_{w,d} = \bmu_w^T\bmv_d$. To maximize $\mathcal{J}_{w,d}$, we should have
\begin{equation}
\small
0 = \frac{\partial \mathcal{J}_{w,d}}{\partial x_{w,d}}=\#_d(w)\sigma(-x_{w,d})-\frac{\#_{\mathcal{D}}(w)\cdot b\cdot \lambda_d}{\lambda_{\mathcal{D}}}\sigma(x_{w,d}).
\end{equation}
That is,
\begin{equation}
\small
e^{2x_{w,d}}-\Big(\frac{\#_d(w)\cdot\lambda_{\mathcal{D}}}{\#_{\mathcal{D}}(w)\cdot b\cdot\lambda_d}-1\Big)e^{x_{w,d}}-\frac{\#_d(w)\cdot\lambda_{\mathcal{D}}}{\#_{\mathcal{D}}(w)\cdot b\cdot\lambda_d} = 0.
\end{equation}
Therefore, $e^{x_{w,d}}=-1$ (which is invalid) or $e^{x_{w,d}}=\frac{\#_d(w)\cdot\lambda_{\mathcal{D}}}{\#_{\mathcal{D}}(w)\cdot b\cdot\lambda_d}$. In other words,
\begin{equation}
\small
\begin{split}
\bmu_w^T\bmv_d &= x_{w,d} = \log\Big(\frac{\#_d(w)\cdot\lambda_{\mathcal{D}}}{\#_{\mathcal{D}}(w)\cdot b\cdot\lambda_d}\Big), \\
\text{\ i.e., \ } \bU_w^T\bV_d &= \bX_{wd}.
\end{split}
\label{eqn:levy1}
\end{equation}

Similarly, for $\mathcal{J}_{\rm category}$, the objective can be rewritten as
\begin{equation}
\small
\begin{split}
& \sum_{w\in \mathcal{V}_{\mathcal{D}}} \sum_{c_i\in \mathcal{C}} \textbf{1}_{w\in \mathcal{S}_i} \log \sigma(\bmu_w^T\bmv_{c_i})\ + \\
& \sum_{w\in \mathcal{V}_{\mathcal{D}}} \textbf{1}_{w\in \mathcal{S}_1 \cup ... \cup \mathcal{S}_{|\mathcal{C}|}} \sum_{c'\in \mathcal{C}}\frac{b}{|\mathcal{C}|}\log \sigma(-\bmu_w^T\bmv_{c'}).
\end{split}
\end{equation}
Following the derivation of $\mathcal{J}_{\rm document}$, we get
\begin{equation}
\small
\begin{split}
\bmu_w^T\bmv_{c_i} &= x_{w,c_i} = \log\Big(\frac{\textbf{1}_{w\in \mathcal{S}_i}|\mathcal{C}|}{\textbf{1}_{w\in \mathcal{S}_1 \cup ... \cup \mathcal{S}_{|\mathcal{C}|}}\cdot b}\Big), \\
\text{\ i.e., \ } \bU_w^T\bV_{c_i} &= \bX_{wc}.
\end{split}
\label{eqn:levy2}
\end{equation}

Putting Eqs. (\ref{eqn:levy0}), (\ref{eqn:levy1}), and (\ref{eqn:levy2}) together gives us Eq. (\ref{eqn:matrix}).

\section{The Ensemble Ranking Function}
\label{sec:app_ensemble}
In Section \ref{sec:ensemble}, we propose the following ensemble ranking function:
\begin{equation}
\small
{\rm score}(w|\mathcal{S}_i) = \bigg(\frac{1}{2}\Big(\frac{1}{{\rm rank}_G(w)}\Big)^\rho + \frac{1}{2}\Big(\frac{1}{{\rm rank}_L(w)}\Big)^\rho\bigg)^{1/\rho}.
\end{equation}
Now we prove this ranking function is a generalization of the \textit{arithmetic} mean reciprocal rank (i.e., MRR) and the \textit{geometric} mean reciprocal rank:
\begin{equation}
\small
\begin{split}
    & \lim_{\rho \rightarrow 1} {\rm score}(w|\mathcal{S}_i) = \frac{1}{2}\Big(\frac{1}{{\rm rank}_G(w)} + \frac{1}{{\rm rank}_L(w)}\Big); \\
    & \lim_{\rho \rightarrow 0} {\rm score}(w|\mathcal{S}_i) = \Big(\frac{1}{{\rm rank}_G(w)} \cdot \frac{1}{{\rm rank}_L(w)}\Big)^{1/2}.
\end{split}
\end{equation}
The case of $\rho \rightarrow 1$ is trivial. When $\rho \rightarrow 0$, we aim to show that
\begin{equation}
\small
    \lim_{\rho \rightarrow 0} \log {\rm score}(w|\mathcal{S}_i) = \log \Big(\frac{1}{{\rm rank}_G(w)} \cdot \frac{1}{{\rm rank}_L(w)}\Big)^{1/2}.
\end{equation}
In fact, let $r_G = \frac{1}{{\rm rank}_G(w)}$ and $r_L = \frac{1}{{\rm rank}_L(w)}$.
\begin{equation}
\small
\begin{split}
    \lim_{\rho \rightarrow 0} \log {\rm score}(w|\mathcal{S}_i)
    & = \lim_{\rho \rightarrow 0} \log \Big(\frac{1}{2}r_G^\rho + \frac{1}{2}r_L^\rho\Big)^{1/\rho} \\
    & = \lim_{\rho \rightarrow 0} \frac{\log \big(\frac{1}{2}r_G^\rho + \frac{1}{2}r_L^\rho\big)}{\rho} \\
    & = \lim_{\rho \rightarrow 0} \frac{\frac{\frac{1}{2}r_G^\rho \log r_G + \frac{1}{2}r_L^\rho \log r_L}{\frac{1}{2}r_G^\rho + \frac{1}{2}r_L^\rho}}{1} \\
    & = \frac{\lim_{\rho \rightarrow 0} \Big(r_G^\rho \log r_G + r_L^\rho \log r_L\Big)}{\lim_{\rho \rightarrow 0} \Big(r_G^\rho + r_L^\rho\Big)} \\
    & = \frac{\log r_G + \log r_L}{2} \\
    & = \log (r_G \cdot r_L)^{1/2}.
\end{split}
\end{equation}
The third line is obtained by applying L'Hopital’s rule.

\end{document}